\documentclass[letterpaper, 10 pt, conference]{ieeeconf}  

\IEEEoverridecommandlockouts                              

\usepackage{algorithm2e}
\usepackage{amsmath,amssymb}
\usepackage{bbm}
\usepackage[english]{babel}
\usepackage[font={small}]{caption}
\usepackage{color}
\usepackage{epic}
\usepackage{epsfig}
\usepackage{epstopdf}
\usepackage[multiple]{footmisc}
\usepackage{graphicx}
\usepackage{hyperref}
\usepackage{placeins}
\usepackage[caption=false,font=footnotesize]{subfig}
\usepackage{wrapfig}
\usepackage[dvipsnames]{xcolor}
\usepackage{mathtools}
\usepackage{fancybox,color}
\usepackage{cite}

\title{\LARGE \bf
Generalized Omega Turn Gait Enables Agile Limbless Robot Turning in Complex Environments
}

\author{Tianyu Wang$^{1,*}$, Baxi Chong$^{1,*}$, Yuelin Deng$^{2}$, Ruijie Fu$^{2}$, Howie Choset$^{2}$, Daniel I. Goldman$^{1}$
\thanks{*These authors contributed equally.}
\thanks{$^{1}$Tianyu Wang, Baxi Chong, and Daniel I. Goldman are with Georgia Institute of Technology, Atlanta, GA 30332, USA.  {\{\tt\small tianyuwang, bchong9\}@gatech.edu, daniel.goldman@physics.gatech.edu}}%
\thanks{$^{2}$Yuelin Deng, Ruijie Fu and Howie Choset are with Carnegie Mellon University, Pittsburgh, PA 15213, USA.  {\{\tt\small yuelinde, ruijief, choset\}@andrew.cmu.edu}}
}

\begin{document}

\maketitle
\thispagestyle{empty}
\pagestyle{empty}


\begin{abstract}

Reorientation (turning in plane) plays a critical role for all robots in any field application, especially those that in confined spaces. While important, reorientation remains a relatively unstudied problem for robots, including limbless mechanisms, often called snake robots. Instead of looking at snakes, we take inspiration from observations of the turning behavior of tiny nematode worms \textit{C. elegans}. 
Our previous work presented an in-place and in-plane turning gait for limbless robots, called an \textit{omega turn}, and prescribed it using a novel two-wave template \cite{wang2020omega}. 
In this work, we advance omega turn-inspired controllers in three aspects: 1) we use geometric methods to vary joint angle amplitudes and forward wave spatial frequency in our turning equation to establish a wide and precise amplitude modulation and frequency modulation on omega turn; 2) we use this new relationship to enable robots with fewer internal degrees of freedom (i.e., fewer joints in the body) to achieve desirable performance, and 3) we apply compliant control methods to this relationship to handle unmodelled effects in the environment.
We experimentally validate our approach on a limbless robot that the omega turn can produce effective and robust turning motion in various types of environments, such as granular media and rock pile.

\end{abstract}


\vspace{-0.3em}
\section{Introduction}
\vspace{-0.3em}
\label{sec:introduction}

Elongate limbless robots are considered promising candidates for tasks in confined environments that other robots and humans cannot access \cite{whitman2018snake, trebuvna2016inspection}.
Previous work on limbless robots has focused on the coordination of their many internal degrees of freedom (DoF) \cite{chong2021frequency,fjerdingen2009snake,takemori2018ladder, lipkin2007differentiable,fu2020robotic} and interactions between their bodies and environments for \textit{\textbf{forward}} locomotion \cite{transeth2008snake,travers2018shape,wang2020directional,fu2020lateral}. 
Less attention has been paid to development of turning motion, which also plays an essential role in applications of limbless robots, especially in the scenarios where high maneuverability is required \cite{liljeback2012snake}. 

\begin{figure}[t]
\centering
\includegraphics[width=0.81\columnwidth]{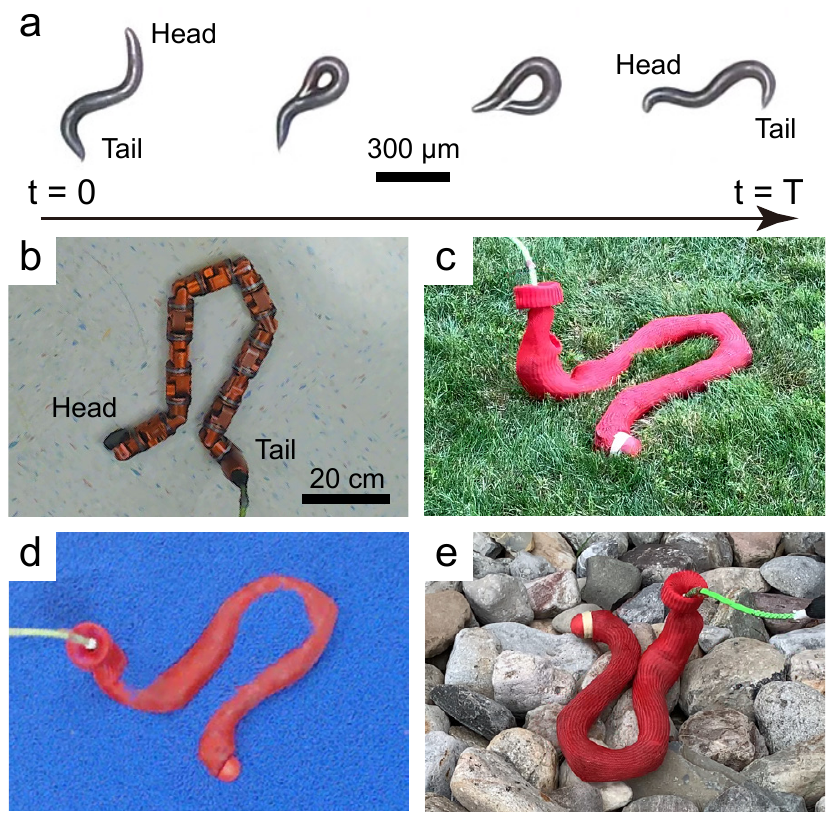}
\caption{The bio-inspired omega turn allows agile limbless robot in-place and in-plane reorientation. (a) The omega ($\Omega$) shaped turning behavior of the nematode worm \textit{C. elegans} in a gait cycle. Limbless robot reorientation on various types of terrain: (b) flat hard ground, (c) rough grassland, (d) granular media, and (e) a pile of rocks.}
\label{fig:intro}
\vspace{-2em}
\end{figure}

Inspired by a turning motion found in the locomotion of a millimeter-scale nematode worm \textit{C. elegans} on heterogeneous terrains \cite{stephens2008dimensionality}, our previous work developed a novel turning gait for limbless robots, the \textit{omega turn}, which drives the robot's ``head" to sweep near the ``tail", inscribing an ``Omega" ($\Omega$) shape for large rotational displacement \cite{wang2020omega}.
Different from commonly used single serpenoid wave gait template \cite{hirose1993biologically}, the omega turn gait design introduced a novel two-wave template\textemdash a superposition of two co-planar traveling waves\textemdash which enables a larger range of body motions. 
Tools found in geometric mechanics \cite{chong2019hierarchical} were to determine parameters which lead to optimal turning performance. 

Through robophysical experiments, the omega turn gait was demonstrated to outperform the widely employed single-wave template turning gaits such as the offset turn (augmenting body waves with a constant offset) \cite{hirose1993biologically,ye2004turning}, and the geometric turn (engineered with techniques from geometric mechanics) \cite{dai2016geometric}; and proved to be a potential candidate for turning motion in confined spaces. 
However, there still are limitations that hold the original implementation of the omega turn back from practical applications, for instance, the increased number of parameters in the template resulted in difficulty of controlling the turning angle; the omega turning performance in dense environments was not robust. 

In this work, we advance the omega turn gait in order to break the aforementioned limitations and expand its applicability. 
Using geometric methods, we vary joint angle amplitude and forward wave spatial frequency in the two-wave equation to establish methods for amplitude and frequency modulation on omega turn, so that the omega turn can produce desired angular displacement and emerge with different body shapes. 
By specifying the relationship between omega wave spatial frequency and the angular displacement, we enable robots with fewer internal degrees of freedom (i.e., fewer joints in the body) to achieve desirable reorientation performance.
By applying the compliant control framework \cite{travers2018shape} during the course of turning motion, we robustify the turning performance in confined environments regardless of the distribution of obstacles. 
We predict turning performance using numerical simulations and verify through robophysical systematic studies. 
By statistical analysis and experimental validation, we show that the advanced omega turn provides effective turning performance on various types of terrain that the present turning strategies cannot reach.


\vspace{-0.3em}
\section{Background and Related Work}
\label{sec:background}

\vspace{-0.5em}
\subsection{Omega Turn}
\vspace{-0.3em}
Inspired by the \textit{C. elegans} turning motion\cite{croll1975components}, our previous work \cite{wang2020omega} developed a template for in-plane turning gaits. 
The template consists of two coplanar traveling sinusoidal waves: a \textit{forward wave} and a \textit{omega wave} (named as \textit{turning wave} in \cite{wang2020omega}).
The template prescribes the joint angles by
\vspace{-0.3em}
\begin{equation}
\begin{aligned}
    \theta_i(t)  =  &A_f(t)\sin{\left(2\pi\omega_f t+2\pi k_f\frac{i}{N}\right)} + \\
    \vspace{-0.7em}
    &A_o(t)\sin{\left(2\pi\omega_o t + 2\pi k_o \frac{i}{N}+\psi\right)},
\end{aligned}
\label{eq:template}
\end{equation}
where $A_{f,o}, \omega_{f,o}$ and $k_{f,o}$ represent the amplitude, the temporal frequency and the spatial frequency for the forward wave and the omega wave, $t$ is the time, $i$ is the index of the joint, $N$ is the total number of joints, and $\psi$ is the phase difference between two waves.
Note that $\omega_f$ is kept the same as $\omega_o$, and phases of the waves are defined as $\tau_f = 2\pi\omega_f t$ and $\tau_o = 2\pi\omega_o t + \psi$.

With the gait template in \eqref{eq:template}, we fixed $k_f = 1.5$ and found omega turning motion emerging at $k_o = 1$, generating effective rotational displacement compared to other turning gaits. 
Thus the gait with $k_o = 1$ in \eqref{eq:template} was denoted by the omega turn for limbless robot.
Using tools from geometric mechanics, we found the optimal phases for two waves in the template that maximize the turning angle of the robot body.
We briefly review the process of optimization using geometric mechanics tools in the next section.

\vspace{-0.5em}
\subsection{Geometric Mechanics}
\vspace{-0.3em}
\begin{figure*}[t]
\centering
\includegraphics[width=0.81\linewidth]{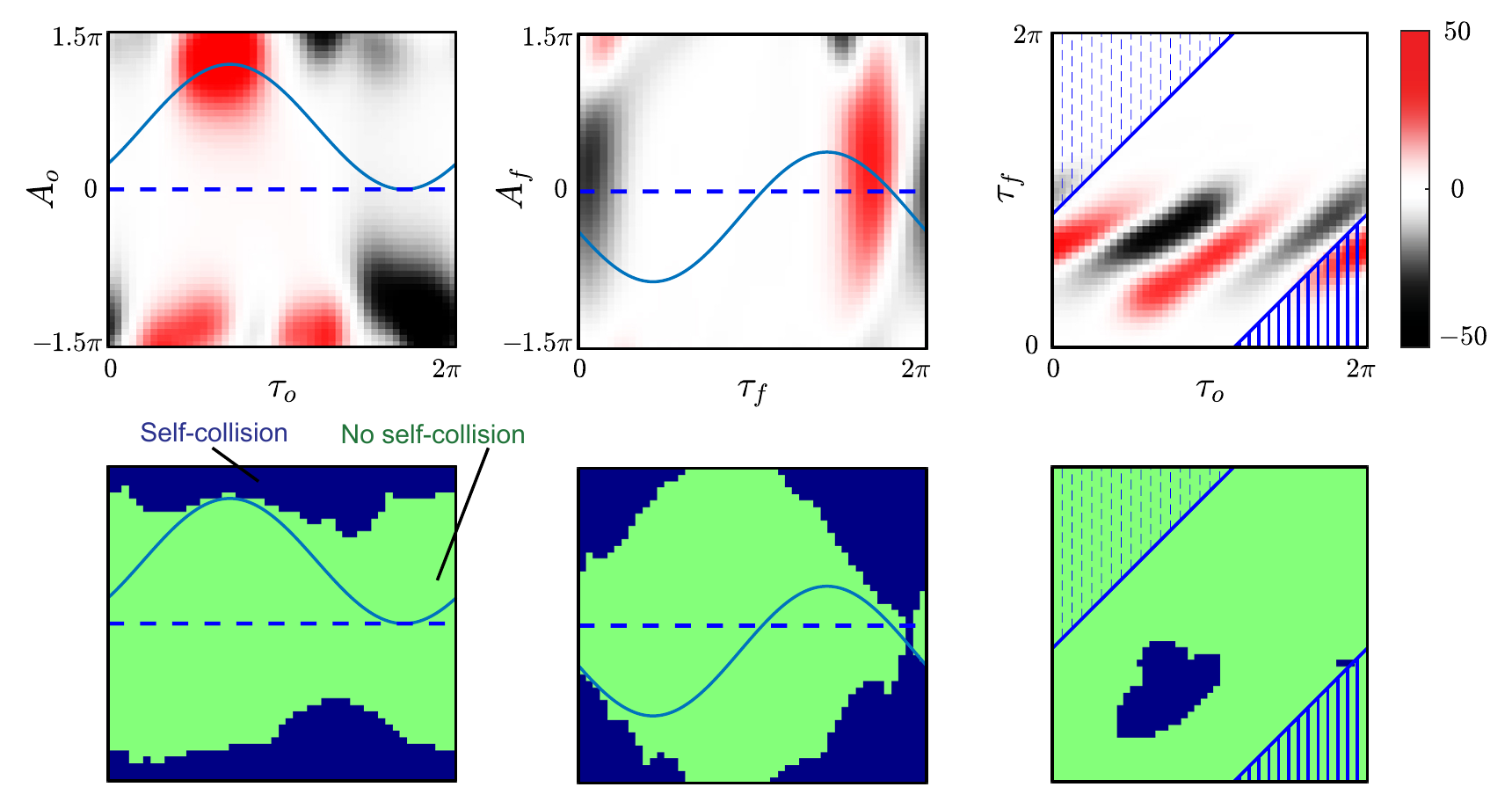}
\vspace{-0.8em}
\caption{The height functions on three 2-dimensional sub-shape spaces. (top) The height function and (bottom) self-collision region on the shape space (a) $\{[\tau_f \ A_f], \tau_f\in S^1, A_f\in \mathbb{R}^1\}$ (b) $\{[\tau_o \ A_o], \tau_o\in S^1, A_o\in \mathbb{R}^1\}$ (c) $\{[\tau_f \ \tau_o], \tau_f\in S^1, \tau_o\in S^1\}$.
The red and black colors represent the positive and negative values of the height function on the top figures. The black regions in the bottom figures represents the shapes that lead to self-collision. The blue curve shows the gait paths $f_1$, $f_2$ and $f_3$, designed to maximize the surface integral while not passing through the collision regions. 
The surface integrals in (a) and (b) is the integral of surface enclosed by the gait path and the dashed line; in (c) is the integral of surface enclosed in the lower right corner (shadow by solid line) minus the surface enclosed in the upper left corner (shadow by dashed line) }
\vspace{-2em}

\label{fig:height_function}
\end{figure*}

We apply the hierarchical geometric framework \cite{chong2019hierarchical} to design omega turn gaits. Specifically, we propose to reconstruct omega turn by superposition of two traveling waves as in \eqref{eq:template}.
By defining $\tau_f = 2\pi\omega_f t$ and $\tau_o = 2\pi\omega_o t + \psi$, we form a four-dimensional shape variable $m = [A_f, \tau_f, A_o, \tau_o]^T$. The set of all shape variables is then defined as $M$. The gait path in the shape space can then be described as: $f: t \mapsto m, t\in S^1, m\in M$. 

Some internal shapes lead to self-collision, which are not desired in robot implementation. Further, as we will discuss later, we modulate the turning by implementing joint angle limit, which also introduces infeasible regions in the shape space. In this way, with the geometry of each module (width: 5 cm, length: 7 cm, N = 8 unless otherwise stated), we construct a feasibility map on the shape space. We thus add the constraints that the gait path of $f$ cannot pass through the infeasible region.

Note that $\tau_f$ and $\tau_o$ are cyclic. In this way, we can simplify the gait path in the four-dimensional shape space to three simple functions in the two-dimensional sub-shape spaces \cite{chong2019hierarchical}: $f_1: \tau_f \mapsto A_f$, $f_2: \tau_o \mapsto A_o$, and $f_3: \tau_f \mapsto \tau_o$ ($f_3^{-1}: \tau_o \mapsto \tau_f$). 
Given any two simple functions, we can reduce the shape space dimension to two. For example, given $f_1$, and $f_3$, \eqref{eq:template} becomes
\vspace{-0.3em}
\begin{equation}
\begin{aligned}
    \theta_i = &f_1 ( f_3 ^{-1}(\tau_o))\sin{\left(f_3 ^{-1}(\tau_o) + 2\pi k_f\frac{i}{N}\right)} \\
               & + A_o\sin{\left(\tau_o + 2\pi k_o\frac{i}{N}\right)} = \theta_i(A_o, \tau_o).
               \label{eq:sub_shape_space}
\end{aligned}
\vspace{-0.3em}
\end{equation}
Given \eqref{eq:sub_shape_space}, we can reduce the original shape space to $\{[\tau_o \ A_o], \tau_o\in S^1, A_o\in \mathbb{R}^1\}$, from which we can numerically calculate the height function to optimize for $f_2$.
Similarly, given $f_1$ and $f_2$, the height functions on $\{[\tau_f \ \tau_o], \tau_f\in S^1, \tau_o\in S^1 \}$ can be numerically calculated; 
given $f_2$ and $f_3$ the height functions on $\{[\tau_f \ A_f], \tau_f\in S^1, A_f\in \mathbb{R}^1\}$ can be numerically calculated. 
In the optimization, we iteratively optimize the three simple functions $f_1$, $f_2$ and $f_3$ until a local maximum in turning angle per gait cycle is reached.

For computational simplicity, we reduce the search space of $f_1$, $f_2$ and $f_3$ by prescribing the functions below:
\vspace{-0.3em}
\begin{equation}
\begin{aligned}
    &f_1:\tau_f \mapsto A_f, A_f = a_f (\gamma+\sin{(\tau_f+\phi_{f})}),\\
    &f_2:\tau_o \mapsto A_o,A_o = A_o (1+\sin{(\tau_o+\phi_o)}).\\
    &f_3:\tau_o \mapsto \tau_f, \tau_f = \tau_o + \psi, 
    \label{eq:prescribe_amp}
\end{aligned}
\end{equation}
The converged height functions and gait paths are shown in Fig. \ref{fig:height_function}. 
It may be possible to obtain slightly higher performing gaits by using more complex functions to describe the trajectory through shape space, e.g. as in \cite{ramasamy2016soap}.
However, simpler functions of paths through the shape space are correspondingly easier to optimize and execute on the robot, while still nearing the performance from such complex functions. 

\vspace{-0.5em}
\subsection{Reference Turning Gaits}
\vspace{-0.3em}

To validate the performance of the advanced omega turn gait in complex environments, we compare it with other commonly employed turning gaits in limbless robots, such as the offset turn \cite{hirose1993biologically,shugen2001analysis}, the geometric turn \cite{dai2016geometric}, and the frequency turn \cite{astley2015modulation,chong2021frequency}.
We give a brief review of these turning gaits that works as reference gaits in this work.

\subsubsection{Offset Turn}
To achieve lateral undulation motion, limbless robots can actuate their joints following the \textit{serpenoid curve} \cite{hirose1993biologically}, an effective single sinusoidal wave template which is identical to the forward wave in \eqref{eq:template}.
A simple way to initiate turning motion while lateral undulating is to add a constant offset $\kappa$ in the body curvature onto the sinusoidal wave, yielding
\vspace{-0.3em}
\begin{equation}
    \theta_i(t) = A\sin\left(2\pi\omega t + 2\pi k \frac{i}{N}\right) + \kappa.
    \label{eq:serpenoid}
\end{equation}
By tuning the offset parameter $\kappa$ in \eqref{eq:serpenoid}, the locomotion direction can be altered. 
Note that the offset turn can be achieved using the two-wave template in \eqref{eq:template} by letting $k_o=0$.

\subsubsection{Geometric Turn}
Dai et al. \cite{dai2016geometric} engineered a turning gait using tools from geometric mechanics. Their work described the serpenoid curve \eqref{eq:serpenoid} as a weighted sum of sine and cosine modes, where weights served as the parameters of the shape space. 
By experimentally determining the local connection \cite{hatton2015nonconservativity} relating trajectories in the shape space to those in the robot position space, a trajectory in the shape space was found to generate maximum displacement in the rotational position space. 
We refer to the gait that realizes the maximum displacement as the ``geometric turn."
Note that the geometric turn can be achieved using the two-wave template in \eqref{eq:template} by letting $k_o = k_f$.

\subsubsection{Frequency Turn}
The frequency turn was developed while studying the sidewinding motion for limbless robots \cite{astley2015modulation,chong2021frequency}. 
In sidewinding motion, two sinusoidal traveling waves were separately implemented in the horizontal and the vertical planes, where the horizontal wave altered locomotion direction and the vertical wave governed the contact between the body and the environment.
By modulating the ratio of spatial frequencies of the two waves, turning motion could be achieved, and was called as a ``frequency turn." 
Compared to the in-plane turning gaits, the frequency turn allows the robot to partially lift its body out of the plane. 


\vspace{-0.5em}
\section{Experiment}
\vspace{-0.3em}
\label{sec:experiment}

Based on the two-wave template as in \eqref{eq:template}, we extend the omega turn gait via parameter modulation and coordination in order to achieve agile turning motion under different conditions and in different environments. For the variants that we will present in Section \ref{sec:results}, we tested the turning performance in numerical simulations and robophysical experiments.

In numerical simulations, under the assumption of quasi-static motion, we determined the instantaneous body velocity from the force and torque balance in ground reaction forces. Then we obtain the body trajectory of the robot in the position space by integrating the body velocity throughout one period \cite{hatton2015nonconservativity,chong2021coordination,chong2021frequency}. We use kinetic Coulomb ground friction \cite{rieser2019geometric} ($F=-\mu\frac{v}{|v|}$, where $F$ is the ground reaction force, and $v$ is the body velocity) to model the ground reaction force on hard ground. Note that under the quasi-static assumption, the net displacement will be independent of the choices of $\mu$.
Then the angular displacement is determined by the orientation change of the averaged main body axis over one gait cycle.

In robophysical experiments, we used a limbless robot composed of 16 identical alternative pitch-yaw arranged rotary joints (unless otherwise specified). 
The gaits were executed by controlling the positions of joints to follow a sequence of joint angle commands. 
Note that for 2D in-plane motion, we only command odd (yaw) joints to move with even (pitch) joint angle maintains zero. 
For each gait tested, we repeated the experiment three times. 
In each trial, we commanded the robot to execute three cycles of the gait.
The motion of the robot was tracked by an OptiTrack motion capture system at a 120 FPS frequency with eight reflective markers affixed along the backbone of the robot.

\vspace{-0.5em}
\section{Results}
\vspace{-0.3em}
\label{sec:results}

We modulated and coordinated the parameters in the omega turn gait template as in \eqref{eq:template} for effective turning motion under various conditions. 
In this section, we present methods and experimental results of the omega turn gait variants, as well as the performance comparisons with the reference turning gaits. 

\vspace{-0.5em}
\subsection{Amplitude Modulation}
\vspace{-0.3em}
\label{subsec:amp_mod}

\begin{figure}[t]
\centering
\includegraphics[width=0.81\columnwidth]{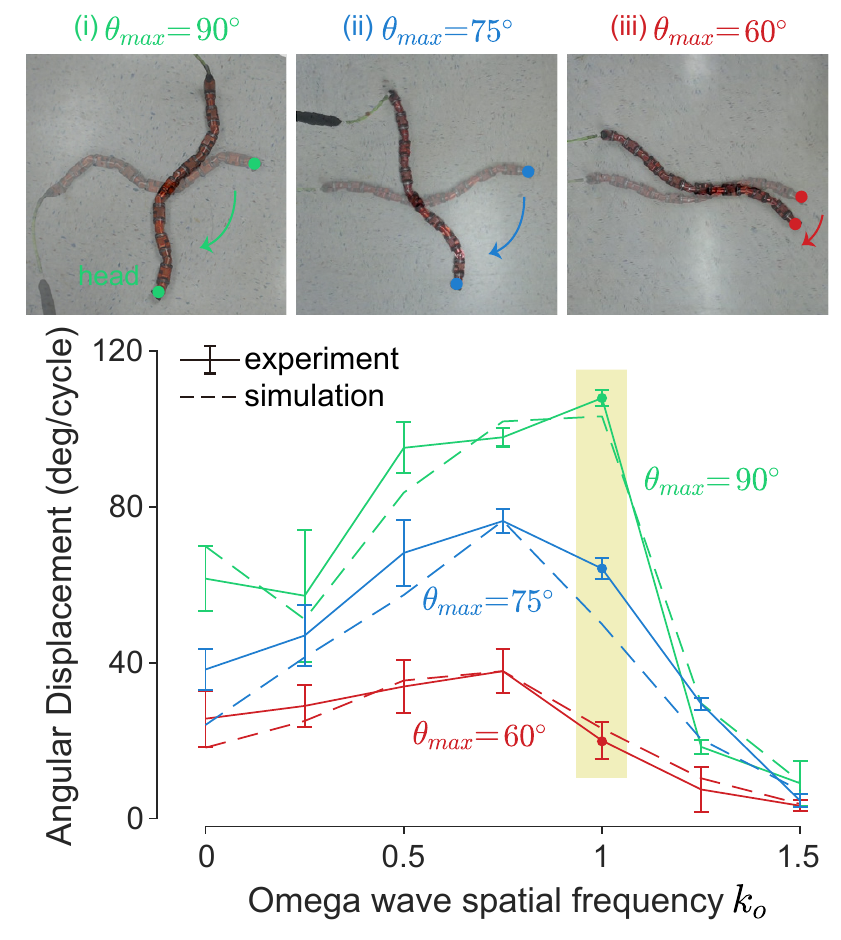}
\caption{Amplitude modulation of turning gaits. The omega turn ($k_o = 1$, highlighted) displays the largest tunable range of angular displacement. Three time-lapse frames of robophysical experiments depicts the courses of turning with joint amplitude $60^\circ, 75^\circ$ and $90^\circ$ in one gait cycle.}
\label{fig:amp_mod}
\vspace{-2em}
\end{figure}

For highly maneuverable limbless locomotion in confined spaces, fine tuning of the locomotive direction is often needed in order to follow a designated path or avoid jamming in between obstacles. 
Thus, it is important to modulate the turning gait to execute exact turning angles in limbless robot agile motion. 
To this end, we varied the parameters two-wave template \eqref{eq:template} to explore a simple way to modulate the rotational displacement of the robot.

We modulated the turning angle by controlling the joint angle limit, $\theta_{max}$. In other words, we define a configuration to be infeasible if $\exists\ i \in \{1\ 2\ ...\ N\}$ such that $|\theta_i| > \theta_{max}$. In this way, we can numerically calculate the infeasible region on each sub-shape space and design gait path to avoid passing through it.

In amplitude modulation experiments, we fixed forward wave spatial frequency $k_f = 1.5$ and tested the gaits with omega wave spatial frequency $k_o$ ranging from 0 to 1.5 on the flat hard ground.
Fig. \ref{fig:amp_mod} depicts the angular displacement per gait cycle as a function of $k_o$ for three different joint angle amplitudes.
The three rows of frames show the time-lapse robot body shapes during the course of turns for different joint angle amplitudes. 
The comparison of turning motion under different joint angle amplitudes can be found in the supplementary video.

The robophysical experiments result shows that the omega turn gait ($k_o = 1$) is capable of producing $20.0^\circ \pm 4.7^\circ$, $64.2^\circ \pm 2.7^\circ$ and $108.0^\circ \pm 2.1^\circ$ of angular displacement per cycle under $\theta_{max} = 60^\circ, 75^\circ$ and $90^\circ$, respectively. 
The modulation of joint angle amplitude between $60^\circ$ and $90^\circ$ can yield a turning angle within the range of $88^\circ$, which is approximately 2 times larger than the range offset turn ($k_o = 0$) and geometric turn ($k_o = 1.5$) can produce ($36.1^\circ$ and $5.8^\circ$). 
This experiment demonstrates that the omega turn gait is capable of generating a continuous range of angular displacement via amplitude modulation, thus is a good candidate for applications in which high maneuverability is required.

\vspace{-0.5em}
\subsection{Spatial Frequency Variation}
\vspace{-0.3em}

\begin{figure}[t]
\centering
\includegraphics[width=0.81\columnwidth]{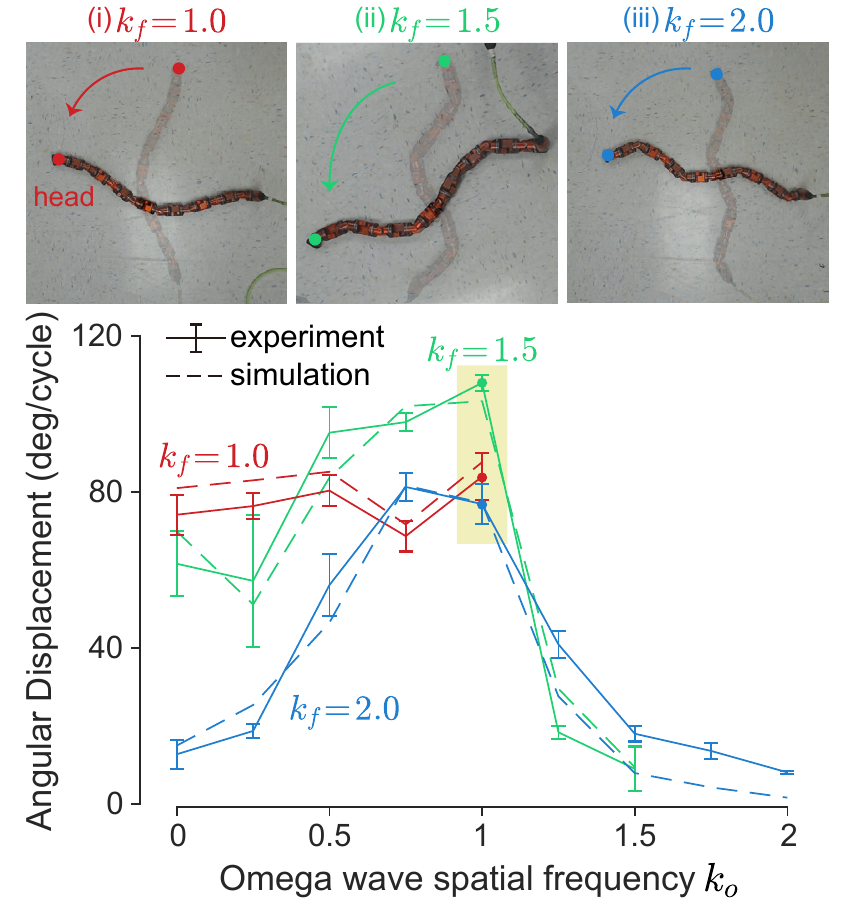}
\caption{Turning gaits with spatial frequency variation. The omega turn ($k_o = 1$) performs robustly over different spatial frequencies of the forward wave $k_f$ (number of waves on the body). Starting and ending positions of the omega turn with varied $k_f$ are shown in the robot pictures.}
\label{fig:sp_freq_mod}
\vspace{-2em}
\end{figure}

During limbless robot locomotion, especially in confined spaces, limbless robots need to frequently vary their body shapes to adapt to the environment. 
Two key parameters that control the body shape in the gait templates that are formed by sinusoidal waves are the amplitude (dictates local body curvature) and the spatial frequency (dictates the number of waves on the body).
As joint amplitude modulation has been discussed in \ref{subsec:amp_mod}, we would like to verify if the omega turn provides consistent turning performance over various forward wave spatial frequencies $k_f$. 

In this set of experiments, we fixed joint angle amplitude $\theta_{max} = 90^\circ$, and tested a series of turning gaits with $k_o \in [0, k_f]$ on the flat hard ground over three forward wave spatial frequencies, $k_f = 1, 1.5$ and $2$. The gaits are designed using the same methods as discussed in previous sections
Fig. \ref{fig:sp_freq_mod} illustrates the simulated and experimental result, while the robot images show the starting and ending positions of omega turns with different $k_f$.  

The result verifies that the omega turn ($k_o = 1$) can provide consistent turning performance over $k_f$: $83.9^\circ \pm 6.1^\circ$, $108.0^\circ \pm 2.1^\circ$ and $76.9^\circ \pm 5.2^\circ$ angular displacement per cycle when $k_f = 1, 1.5$ and $2$, respectively. 
Given the offset and the geometric turning gaits cannot maintain consistent turning performance, the omega turn is a better turning strategy that can be employed conveniently in tasks when the limbless robot is operated with varying body shapes.
Note that the omega turn and the geometric turn are identical when $k_f = 1$.

\vspace{-0.5em}
\subsection{Omega Turn with Different Internal DoF}
\vspace{-0.3em}

\begin{figure}[t]
\centering
\includegraphics[width=0.81\columnwidth]{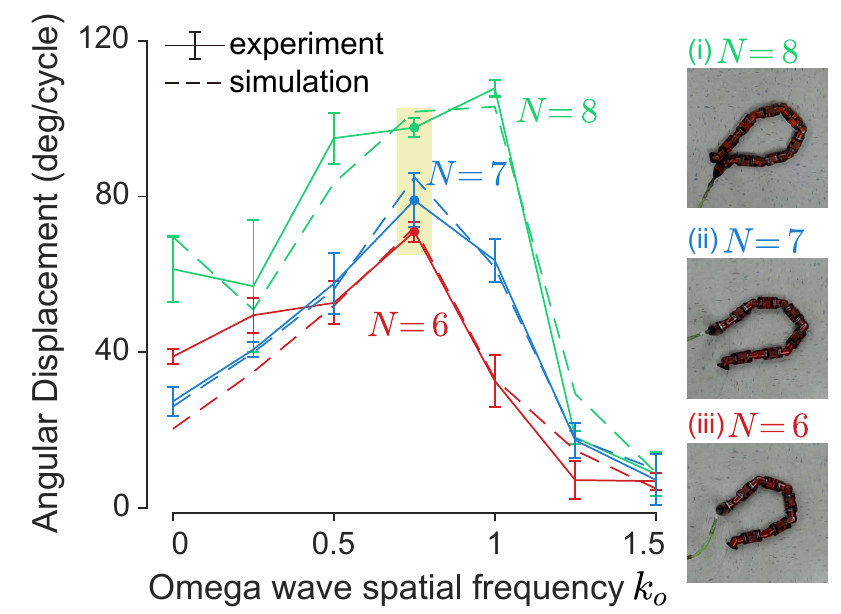}
\caption{Performance of turning gaits on limbless robots with varied internal DoF (joint number). The omega turn can be generalized to different body lengths with fine tuning of omega wave spatial frequency, as the local maximum of angular displacement shifts to $k_o=0.75$ as the joint number decreases. Robot pictures show the key frames when the robot has the largest local body curvature to form the ``$\Omega$" shape.}
\label{fig:joint_mod}
\vspace{-2em}
\end{figure}

Based on needs of the task and constraints created by the environment, limbless robots are used with varying number of rotary joints. 
To expand the applicability of the omega turn strategy, we explored the omega turn for limbless robots with different numbers of joints.

We tailored the omega turn gait for different number of joints ($N$) and tested them on three limbless robots with 6, 7 and 8 of yaw joints. 
Fig. \ref{fig:joint_mod} shows the result of turning performance for the family of gaits with $k_o \in [0, 1.5]$ with $k_f$ fixed at 1.5 and $\theta_{max}$ fixed at $90^\circ$. The gaits are designed using the same methods as discussed in previous sections.
The turning performance of the omega turn ($k_o = 1$) drops modestly as number of joints decreases, and local maximum of turning performance shifts to $k_o = 0.75$ for the cases of $N = 6$ ($71.2^\circ \pm 2.6^\circ$) and $N=7$ ($79.3^\circ \pm 6.9^\circ$).
We posit that this shift of local maximum results from that the head of the robot is no longer able to touch the tail with a shorter body length.
For the robots with shorter body length, gaits with $k_o = 0.75$ allow a shorter distance between head and tail during the course of turn, enable a larger local body curvature, and thus larger angular displacement. 
When $N=8$, $k_o = 0.75$ and $k_o = 1$ both ensure the head to touch the tail when turning. 
The result implies that, the omega turn gait is applicable to a wide range of limbless robots with varied body lengths through an alternation in the omega wave spatial frequency $k_o$, and offers improved turning performance compared to reference gaits. 

\vspace{-0.5em}
\subsection{Omega Turn in Granular Media}
\vspace{-0.3em}

Limbless locomotion is not only employed on hard surfaces, but also has been demonstrated to be useful in granular substrates such as sand \cite{marvi2014sidewinding}. 
Therefore, we studied the turning motion in granular media using a test pool filled with 6mm plastic spheres.

\begin{figure}[t]
\centering
\includegraphics[width=0.81\columnwidth]{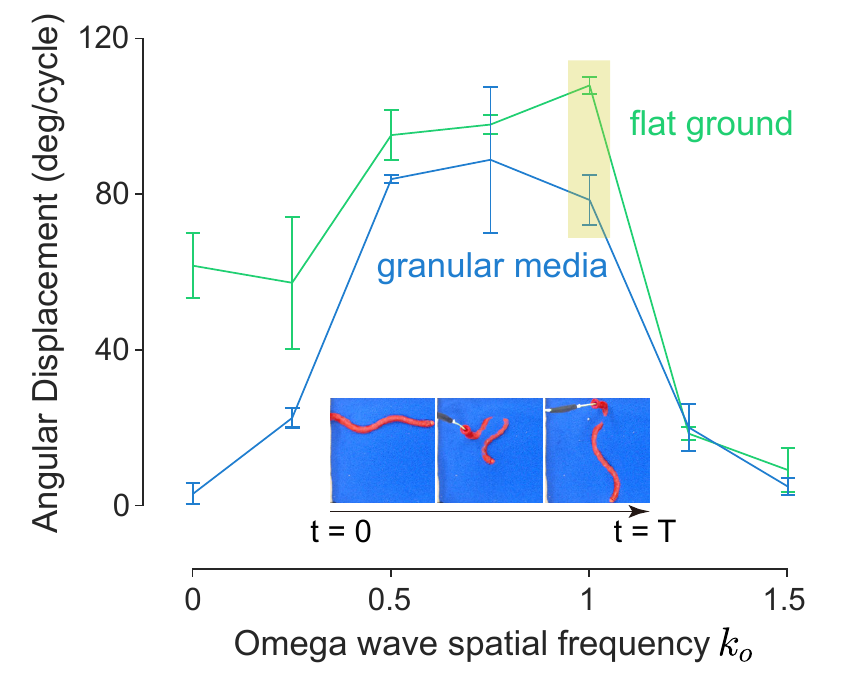}
\caption{Turning gaits in granular media. The omega turn ($k_o = 1$) produces the angular displacement that approaches that on the flat ground. A series of robot pictures show the course of omega turning in granular media.}
\label{fig:granular}
\vspace{-2em}
\end{figure}

We tested a series of turning gaits on the surface of granular media.
Fig. \ref{fig:granular} depicts experiment data and a series of time-lapse key frames for the omega turn in granular media. 
In granular media, the omega turn ($k_o = 1$) can generate $78.5^\circ \pm 6.5^\circ$ angular displacement per cycle, while the offset turn and the geometric turn were ineffective ($2.9^\circ \pm 2.7^\circ$ and $4.8^\circ \pm 2.1^\circ$). 
This comparison indicates that, in granular media, the omega turn is capable of producing effective turning motion which is comparable to the performance on the flat hard ground, while other common turning strategies do not work well.

\vspace{-0.5em}
\subsection{Compliant Omega Turn}
\vspace{-0.3em}
\label{subsec:compliant}

Interacting the obstacles with proprioceptive torque sensors and deforming the body shape to comply to obstacles have been studied for limbless robot forward locomotion \cite{travers2018shape,wang2020directional}. 
Our preliminary results of omega turn in obstacle-rich environments \cite{wang2020omega} demonstrated that, although the omega turn is a promising candidate for tuning in confined spaces, the turning performance is sensitive to initial position and the distribution of surrounding obstacles, since the robot can become wedged between obstacles. 
We hypothesised that the application of the compliant control framework \cite{travers2018shape} on the omega turn motion could enable the robot to compliantly negotiate obstacles during the course of turning.

As an extension of admittance control \cite{murray2017mathematical} to articulated locomotion, the compliant control framework for limbless locomotion assigns spring-mass-damper-like dynamics to the shape parameters in the gait equation to allow them to vary according to the sensed joint torques. 
In this work, we built the compliant control system on wave amplitudes $A = [A_f, A_o]^T$ in the two-wave template \eqref{eq:template} by
\begin{equation}
    M \ddot{A} + B \dot{A} + K (A - A_0) = J\tau_{\text{ext}},
    \label{eq:admittance}
\end{equation}
where $M = \begin{bmatrix} 1 & 0\\ 0 & 1 \end{bmatrix}$, $B = \begin{bmatrix} 8 & 0\\ 0 & 8 \end{bmatrix}$, and $K = \begin{bmatrix} 8 & 0\\ 0 & 8 \end{bmatrix}$ are positive-definite tuning matrices ($2\times 2$) that govern the dynamic response, $A_0 = [45^\circ, 45^\circ]^T$ is the nominal amplitude, $\tau_{\text{ext}}$ is the vector of external torques ($N\times 1$) measured by the joint torque sensors, and $J = [\sin(2\pi\omega_ft+2\pi k_f\frac{i}{N}), \sin(2\pi\omega_o t+2\pi k_o\frac{i}{N})]^T$ is the Jacobian ($2\times N$) that maps the external torques onto the amplitude. 
We solved \eqref{eq:admittance} for $\ddot{A}$ and double integrated $\ddot{A}$ using Newton-Euler method for the amplitude. 
We refer readers to \cite{travers2018shape} for detailed explanation and demonstration of the compliant control framework on limbless robot locomotion. 

\begin{figure}[t]
\centering
\includegraphics[width=0.81\columnwidth]{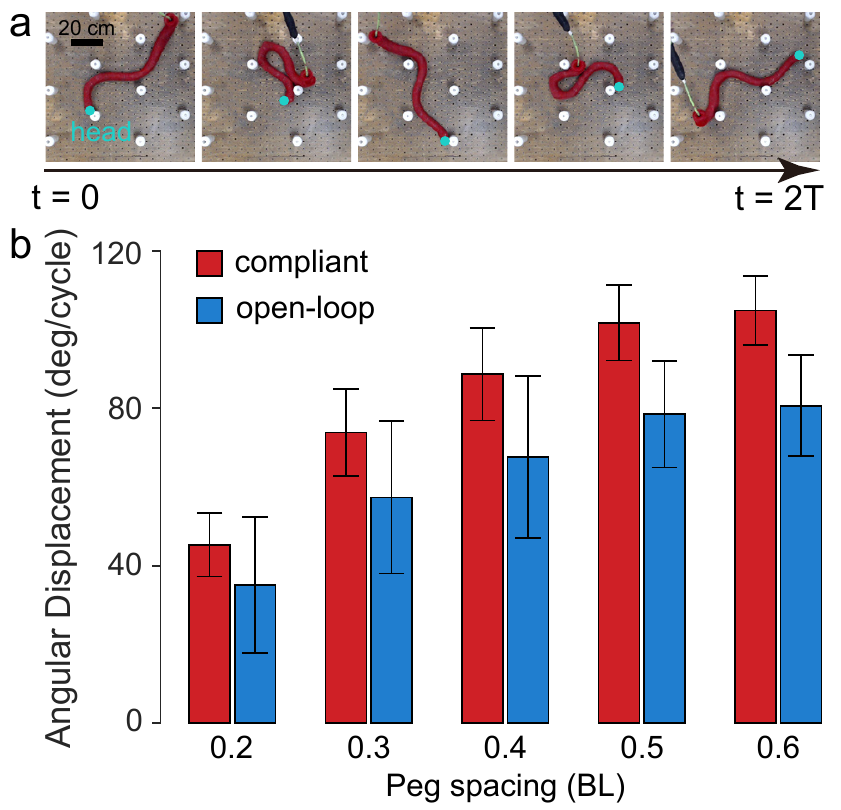}
\caption{The omega turn with the compliant control applied on the peg board with varied peg spacing (in body lengths, BL). (a) Time-lapse images of a limbless robot executing the compliant omega turn in a peg board with 0.3 BL spacing. (b) The compliant omega turn generates larger averaged turning angle compared to the open-loop turn, as well as performs more consistent (shorter error bar). }
\label{fig:compliant}
\vspace{-2em}
\end{figure}

We tested the compliant omega turn on an indoor artificial obstacle-rich environment\textemdash a board with a hexagonal array of pegs, and compared with open-loop omega turn in the same environment.
Fig. \ref{fig:compliant}a illustrates the comparison of turning performance between open-loop and compliant omega turns, where the angular displacement increased by $\sim$30\% with the compliant omega turn in all tested environments with various peg spacing ranging from 0.2 body lengths (BL) to 0.6 BL.
Furthermore, the compliant omega turn performed more robustly with smaller variance in angular displacement in all environments, reflected by shorter error bars (standard deviation) than the open-loop variant. 
Fig. \ref{fig:compliant}b captures some key body shapes during two cycles of turning in a peg board with 0.3 BL spacing. 
An example of compliant omega turn in peg boards can also be found in the supplementary video.

\vspace{-0.5em}
\subsection{Omega Turn on Complex Terrain}
\vspace{-0.3em}

Finally, to test if our lab robophysical studies could show benefit in field robot application, we carried out field experiment by running the omega turn gait developed in Section \ref{subsec:compliant} on an outdoor pile of rocks (diameter $\sim$0.3 BL) where distribution of obstacles and contact between body and environment are nondeterministic. 
We also tested the offset turn gait, the geometric turn gait, and the frequency turn gait \cite{marvi2014sidewinding} as references. 

\begin{figure}[t]
\centering
\includegraphics[width=0.81\columnwidth]{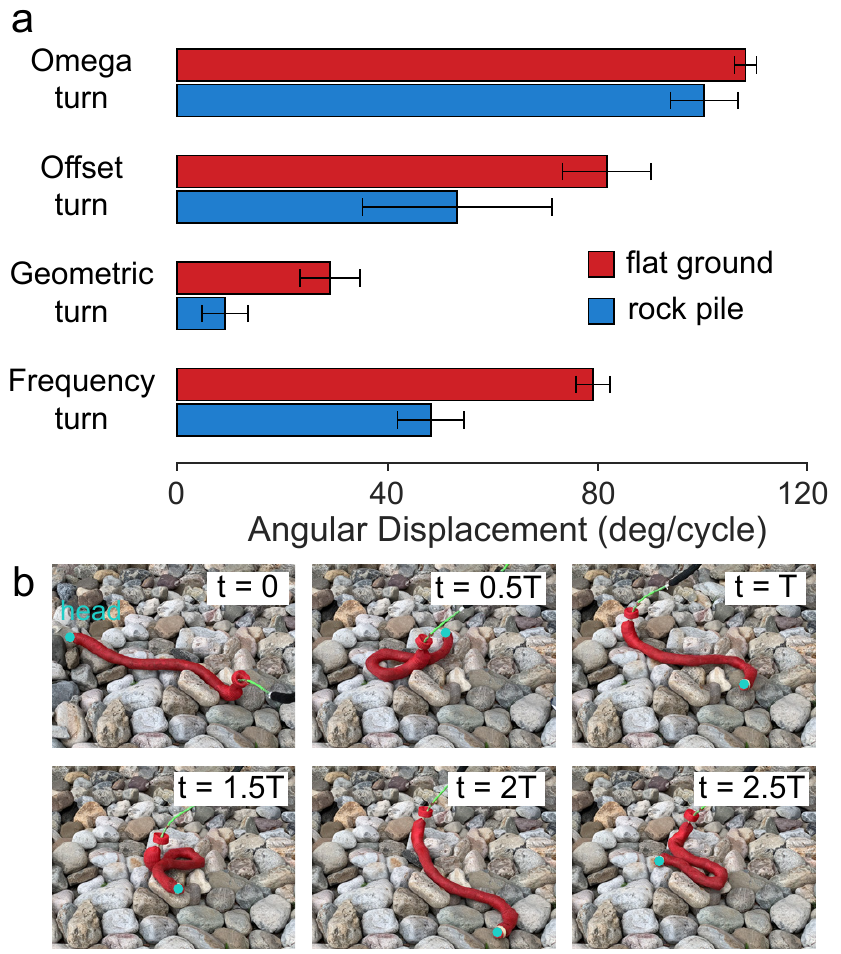}
\caption{Comparison of different turning strategies on hard ground and on an outdoor rock pile. (a) The omega turn outperforms other common turning strategies in both environments, and its performance on the rock pile ($100.1^\circ \pm 6.4^\circ$) approaches that on hard ground ($108.0^\circ \pm 2.1^\circ$). (b) Time-lapse frames show the omega turn enables agile reorientation of a limbless robot on the rock pile.}
\label{fig:rock}
\vspace{-2em}
\end{figure}

Fig. \ref{fig:rock}a presents the averaged angular displacement on the rock pile for each of tested gaits, as well as their turning performance on the flat hard ground.
Although performance of all the gaits drops when executed on the rock pile, the omega turn was still capable of generating a ${\sim}100^\circ$ of angular displacement per cycle.
Also, the performance of the omega turn was robust on the rock pile, given a small standard deviation of $6.4^\circ$.
Selected key frames of the limbless robot executing an omega turn on the rock pile are presented in Fig. \ref{fig:rock}b, and the whole course of it can be found in the supplementary video.

This set of experiments demonstrated that, with proper modulation and coordination of parameters in the gait template, the omega turn is able to outperform other turning strategies, which makes it a promising approach for effective and robust turning motion in agile limbless locomotion in complex environments. 

\vspace{-0.3em}
\section{Conclusion}
\label{sec:conclusion}
\vspace{-0.3em}

Based on the preliminary work on the development of a bio-inspired omega turn gait for limbless robot \cite{wang2020omega}, this work studied in detail the performance of omega turn gait with variation of key parameters in its template, and extended applicability of the omega turn to more practical scenarios which might be encountered by robot in field. 
1) By modulating the amplitude of joint angles, the omega turn produced the desired angular displacement.
2) By modulating the forward wave spatial frequency, omega turn performed consistently over various forward wave spatial frequencies.
We experimentally demonstrated that the omega turn is achievable by a diverse range limbless robots with varied internal degrees of freedom.
3) Through granular media experiment, the omega turn was verified to generate the largest angular displacement compared to other turning strategies such as the offset turn and the geometric turn.
4) To deal with the obstacles in confined spaces, we introduced the compliant control framework to the omega turn, which allows the robot to comply to obstacles during turning and offered a $\sim$30\% increase of performance. 
5) Finally, field experiments were carried out on an outdoor rock pile to test the performance of different turning strategies in nondeterministic environments. 
Experimental result validated that the omega turn displayed the most effective and robust performance on the rock pile which approached its performance on the flat hard ground. 

This work suggests that proper modulation and coordination of parameters make the omega turn a promising candidate for agile turning motion in limbless robot locomotion.
Future work will study the omega turn motion in more challenging environments.
A next step for our research is to develop transitions between the omega turn gait and forward motion gaits, so that the omega turn can be used in higher level motion planning for practical tasks in diverse applications.




\bibliographystyle{IEEEtran}
\bibliography{OmegaTurn}

\end{document}